\documentclass[10pt,twocolumn,letterpaper]{article}

\usepackage{cvpr}              

\usepackage{graphicx}
\usepackage{amsmath}
\usepackage{amssymb}
\usepackage{booktabs}

\usepackage{balance}

\usepackage[pagebackref,breaklinks,colorlinks]{hyperref}

\usepackage[toc,page,titletoc]{appendix}
\usepackage[capitalize]{cleveref}
\crefname{section}{Sec.}{Secs.}
\Crefname{section}{Section}{Sections}
\Crefname{table}{Table}{Tables}
\crefname{table}{Tab.}{Tabs.}
\crefname{sup}{Supplement}{Supplements}

\usepackage[table]{xcolor}
\definecolor{lightgray}{gray}{0.85}

\def\cvprPaperID{6674} 
\def\confName{CVPR}
\def\confYear{2022}

\begin{document}

\title{Convolutions for Spatial Interaction Modeling}

\author{Zhaoen Su, Chao Wang, David Bradley, Carlos Vallespi-Gonzalez \and Carl Wellington, Nemanja Djuric\\
{\tt\small \{suzhaoen, chao.wang, dbradley, cvallespi, cwellington, ndjuric\}@aurora.tech}
}
\maketitle

\begin{abstract}
In many different fields interactions between objects play a critical role in determining their behavior. Graph neural networks (GNNs) have emerged as a powerful tool for modeling interactions, although often at the cost of adding considerable complexity and latency. In this paper, we consider the problem of spatial interaction modeling in the context of predicting the motion of actors around autonomous vehicles, and investigate alternatives to GNNs. We revisit 2D convolutions and show that they can demonstrate comparable performance to graph networks in modeling spatial interactions with lower latency, thus providing an effective and efficient alternative in time-critical systems. Moreover, we propose a novel interaction loss to further improve the interaction modeling of the considered methods.

\end{abstract}

\section{INTRODUCTION} 
\label{sect:introduction}

Interactions or relations between objects are critical for understanding both the individual behaviors and collective properties of many systems. 
Conceptually, these interactions can be modeled with graph structures that comprise a set of objects (nodes) and their relationships (edges). 
By applying deep learning techniques, graph neural networks (GNNs) have demonstrated great expressive power in modeling interactions in various fields, including physical science \cite{battaglia2016interaction, fout2017protein, sanchez2018graph, Qasim_2019}, social science~\cite{kipf2016semi, hamilton2017inductive}, knowledge graphs~\cite{hamaguchi2017knowledge}, and other research areas~\cite{khalil2017learning, lowe2020amortized, meirom2020stop, xu2020inductive}. 

Some of the interactions strongly depend on geometries, such as the Euclidean distance and relative directions between objects, which in this work is referred to as spatial interactions. One problem where spatial interaction is critical is motion forecasting, a key task in the fields of computer vision, robotics in general and autonomous driving (AD) in particular. Specifically, anticipating the future movements of an object requires understanding not only its history, but also the object's interactions with other objects and its environment. These interactions strongly depend on relative spatial features between objects, such as their relative locations, orientations, and velocities.

Graphs have achieved success in modeling spatial interaction~\cite{schlichtkrull2017modeling, sadeghian2017car-net, kipf2018neural, Rhinehart2019PRECOG}. 
Features of individual objects are typically encoded into attributes of graph nodes, and the graph edges are built by passing node attributes and the relative geometries of the node pair through a mapping function. 
GNNs follow a message passing scheme, where each node aggregates features of its neighboring nodes to compute its new node attributes.  
These approaches have two characteristics, as seen in the experimental section: (1) the relative spatial features are not represented implicitly in the graph and need to be handcrafted into the graph edge features; (2) even a single iteration of GNN may be slower than convolutional neural networks (CNNs), which makes GNNs less suitable for applications in fields such as AD where fast inference is safety-critical.

Alternatively, data structures for $2$D or $3$D convolutional operations are presented in common grid forms, such as through voxelization in 3D, rasterization in 2D bird's-eye view (BEV), or as intermediate CNN features. 
Importantly, spatial relations are intrinsically represented in these Euclidean space. 
Thus, they theoretically allow spatial relations between objects to be learned by CNNs with sufficiently large receptive fields~\cite{engelmann2020dilated}.  
In other words, CNNs have the potential to model spatial interactions. 
However, even though deep CNN backbones with large receptive fields are widely utilized in trajectory forecasting models, research has shown that adding a GNN after a CNN backbone can still improve interaction modeling \cite{sadeghian2017car-net, casas2019spatially, Rhinehart2019PRECOG}.  
This suggests the CNN backbones often do not fulfill their theoretical potential in modeling spatial interactions between the traffic actors.


In this work, we consider spatial interaction modeling through $2$D convolutions and compare them to GNNs within the context of motion forecasting for AD.
A key determinant of future motion for other drivers is the avoidance of collisions, which represents a critical interaction that we model explicitly.
Collisions can be approximated as geometric overlapping, which provides unambiguous definitions for interaction metrics. 
We evaluate the methods on large-scale real-world AD data to draw general conclusions. 
Our contributions are summarized below:
\begin{itemize}
  \item we identify three components to facilitate modeling spatial interaction with convolutions: (1) large actor-centric interaction region, (2) projecting feature maps into the actor's frame of reference, and (3) aggregation of per-actor feature maps using convolutions;
  \item we perform empirical studies to compare interaction modeling using convolutions and graphs, and find that (1) CNNs can perform similarly to or better than GNNs; (2) adding the CNN can considerably improve interaction modeling even when a GNN is used; (3) adding a GNN demonstrates only minor additional gain when the convolutional approach is already used.
  \item we propose and study a novel interaction loss.
\end{itemize}
\section{RELATED WORK}
\label{sect:related_work}

\subsection{Motion forecasting}
There exists a significant body of work on forecasting the motion of traffic actors.
An input to the forecasting models can be a sequence of past actor states such as positions, headings, or velocities \cite{dp2018, boris2018Modeling, Deo2018Conv, Rhinehart2019PRECOG, Diehl2021graph, gao2020vectornet}, or a sequence of raw sensor data such as LiDAR or radar returns \cite{casas2018intentnet, luo2018fast} where joint object detection and motion forecasting are performed in an autonomous vehicle's (AV) frame of reference.
While the latter approach may accelerate inference and joint learning by sharing common CNN features among all actors, these single-stage models could benefit from actor-centric features. 
Two-stage models \cite{casas2019spatially,djuric2020multixnet} address this issue by using a first stage to detect the actors and extract features, and then adding a second stage in the frame of reference of detected actors. 
The two stages are then learned jointly in an end-to-end fashion. 
The interaction modeling study in this paper adapts a two-stage architecture.
Note that the designs used in the study, including rotated region of interest (RROI) \cite{ma2017rroi} and actor-centric design \cite{dp2018, casas2019spatially, djuric2020multixnet}, have been developed and applied in previous research in a context different from interaction modeling.
However, our empirical study demonstrates that utilizing these ideas allows convolutions to effectively model spatial interaction as well. 

\subsection{Interaction modeling}
GNNs have recently been applied to explicitly express interactions in motion forecasting. 
NRI \cite{kipf2018neural} models the interaction between actors by using GNNs to infer interactions while simultaneously learning dynamics. 
VectorNet~\cite{gao2020vectornet} and CAR-Net~\cite{sadeghian2017car-net} model actor-context interactions. 
Closely related to our work, SpaGNN \cite{casas2019spatially} is also a two-stage detection-and-forecasting model that builds a graph for vehicles in the second stage to model vehicle-vehicle interaction. 
The GNN models used for comparison in the study of this paper follow the same design. 

Beyond graph models, grid-based spatial relations have been explored using social pooling approaches~\cite{alahi2016social,Deo2018Conv, gupta2018social-gan}, where pooling is used to capture the impact of surrounding actors in the recurrent architecture. 
In social-LSTM \cite{alahi2016social, gupta2018social-gan}, the LSTM cell receives pooled spatial hidden states from the LSTM cells of neighbors that are embedded into a grid.  
Besides the parameter-free pooling, convolutional layers have also been explored~\cite{Deo2018Conv}. 
By contrast, our proposal is fully convolutional. Moreover, these approaches pool the spatial context of interacting actors while excluding the actor itself, thus the actor-context interaction is not directly modeled in the process. 


\subsection{Interaction metrics}
It is interesting to note that while various techniques have been developed to model spatial interaction, most prior work reports motion forecasting displacement errors. 
As shown in this study, reducing displacement errors does not necessarily indicate improvement in interaction modeling for a motion forecasting task. 
An alternative metric that can more explicitly indicate the level of interaction modeling is to measure whether vehicle motion forecasts incorrectly predict overlap with other vehicles~\cite{casas2019spatially, Rhinehart2019PRECOG}. 
In this work, we also propose vehicle-obstacle overlap rate within motion forecasts as another measure for interaction modeling. 

\begin{figure*}[h!]
    \centering
    \includegraphics[width=0.85\textwidth]{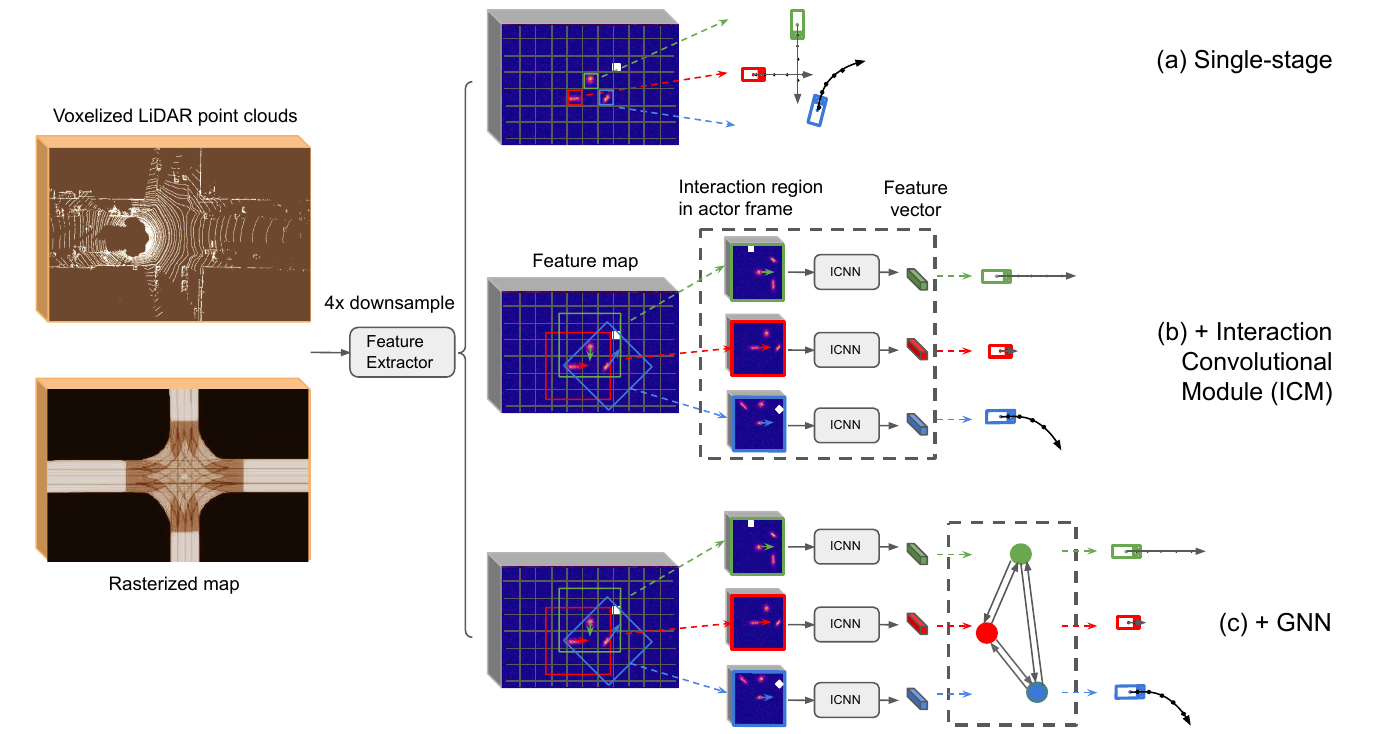} 
    \caption{Three model architectures in a scene illustrated with three vehicle actors and one obstacle (denoted by the white spot). All models share the same first-stage design shown from left to middle: input is a BEV raster image comprising past and current point clouds and a semantic map in the AV frame. Through a CNN feature extractor we obtain a 4$\times$ downsampled feature map in the AV frame. ({\bf a}) Single-stage baseline: Object detection and trajectory forecasting are performed at a pixel-level. ({\bf b}) Adding the proposed Interaction Convolutional Module (ICM). For each actor we define an interaction region (IR) in the actor frame that is used to crop an area from the feature map. Through the weight-sharing interactive CNN (ICNN) a feature vector is aggregated for each actor and then utilized to predict a future trajectory in its frame. ({\bf c}) Adding GNN into the architecture shown in (b).}
    \label{fig:architecture}
\end{figure*}

\section{METHODOLOGY}
\label{sect:methodology}
In this section we formulate the motion forecasting problem, followed by a discussion of two approaches to interaction modeling: implicitly through $2$D convolutions and explicitly through graphs.
Fig.~\ref{fig:architecture} illustrates the architectures of the considered end-to-end models that jointly solve tasks of object detection and motion forecasting, taking BEV representation of the sensor data as an input and outputting both object detections and their future trajectories. 
We emphasize that we purposefully choose a commonly used input representation, neural network design, and loss functions in order to focus on understanding the interaction modeling aspect of these approaches. 
Moreover, to quantify the analysis we limit our discussion to vehicle actors (see Appendix for analysis of other actor types).


\subsection{Problem formulation}
Given input data comprising the past and current information of $V$ interacting actors and the environment, a model outputs their current and future states ${\bf x}$ represented as $\mathcal{X}_{0:H} = \{{\bf x}^{v}_t, v = 1,\ldots,V, ~t = T_0,\ldots,T_H\}$.
As mentioned previously, our study considers raw sensor data as an input to the model. 
Following the joint detection and forecasting architecture \cite{casas2018intentnet, djuric2020multixnet}, we encode the sensor data by voxelizing and stacking a sequence of current and $P$ past LiDAR point clouds around AV at time $T_0$ in BEV representation, as well as rasterizing semantic map that provides an additional environmental prior, which are used as the model input. 
The $2$D detection at time $T_0$ for each actor is parameterized by a bounding box represented as $({c}_{x}, {c}_{y}, \cos\theta, \sin\theta, w, l)$, denoting the $x$ and $y$ coordinates of the actor's centroid, the cosine and sine of its heading angle, and the width and length of the box, respectively. 
Assuming rigid SE2 transformations, future trajectories can be represented as a sequence of tuples $({c}_{xt}, {c}_{yt}, \cos\theta_t, \sin\theta_t)$, with $t \in \{T_1, \ldots, T_H\}$ \cite{su2020temporallycontinuous}. 

\subsection{Feature extraction and loss functions}
As illustrated in Fig.~\ref{fig:architecture}a, the first stage of the joint model detects objects and extracts features. 
From the input BEV raster, a 4$\times$ downsampled feature map is extracted by a deep CNN that follows common design (see Appendix for the complete network design). 
It consists of 3 operations: (1) convolutional block ({ConvB}) including a convolution (kernel size 3$\times$3), batch normalization, and ReLU optionally; (2) ResNet v2 block ({ResB})~\cite{he2016identity}; and (3) upsampling using bi-linear interpolation. 
Features are processed at multiple scales to provide larger receptive fields for capturing wider context and past motion of the actors.

Following the computation of the BEV feature map, classification and regression are performed on the 1D feature vector for each grid cell.  
Through a fully-connected (FC) layer and a softmax function, we obtain the likelihood $p_c$ of existence of a vehicle actor whose center is located in the cell $c$. 
We use focal loss $\ell_f$ \cite{lin2017focalloss} to address the foreground/background imbalance. 
Through a separate FC layer, the network at the same time regresses the detection bounding boxes $\mathcal{X}_{0}$. The centroid and heading are relative to the cell center and the AV heading, respectively.
Then, the first-stage detection loss is given as follows (the hat-notation ${\hat \ast}$ indicates the ground-truth targets)
\begin{align}
\label{eq:first_stage_loss}
\begin{split}
    \mathcal{L}_{det} = ~&  \sum_{c \in \mbox{\footnotesize {$all$}}}\ell_f(\hat p_c, ~p_c) + \sum_{v \in \mbox{\footnotesize $veh$}} \Big( \ell_1(\hat l^v - l^v)  \\
    &  + \ell_1(\hat w^v - w^v)+ \ell_1(\hat{c}_{x0}^v - c_{x0}^v) + \ell_1(\hat{c}_{y0}^v - c_{y0}^v) \\ & + \ell_1(\cos\hat{\theta}_0^v - \cos\theta_0^v) + \ell_1(\sin\hat{\theta}_0^v - \sin\theta_0^v) \Big),
\end{split}
\end{align}
where $all$ and $veh$ represent all grid cells and vehicle foreground grid cells, respectively, $\hat p_c$ equals 1 for foreground cells and 0 otherwise, $\ell_1$ is smooth-$L_1$ loss (with the transition value set to 0.1).

In addition to the detection loss, end-to-end models also optimize for the prediction loss that is only applied to future waypoints of the actors. 
Moreover, we model the multimodality of the predictions~\cite{djuric2020multixnet} by classifying three modes for each actor (i.e., turning left, turning right, or going straight), where a separate trajectory is regressed for each mode along with the corresponding mode probability \cite{cui2019multimodal} based on the focal loss. 
In addition, regression loss is applied only to the trajectory mode that is closest to the observed trajectory. 
Then, the prediction loss is given as
\begin{align}
\begin{split}
    \mathcal{L}_{pred} = ~&  \sum_{v \in \mbox{\footnotesize {$veh$}}} \sum_{m=1}^{M=3} \Big( \ell_f(\hat p_m^v, ~p_m^v) \\
    & + \frac{1_{{\footnotesize \hat{m}=m}}}{H} \sum_{t=T_1}^{T_H} \big(\ell_1(\hat{c}_{xt}^v - c_{xmt}^v) + \ell_1(\hat{c}_{yt}^v - c_{ymt}^v) \\ & + \ell_1(\cos\hat{\theta}_t^v - \cos\theta_{mt}^v) + \ell_1(\sin\hat{\theta}_t^v - \sin\theta_{mt}^v) \big) \Big),
\end{split}
\end{align}
where ${p_m^v}$ denotes probability of $m$-th trajectory mode of actor $v$, $1_c$ is an indicator function equaling $1$ if the condition $c$ holds and $0$ otherwise, and $\hat{m}$ indicates an index of the mode closest to the ground truth. 
Future centroids and headings are relative to the cell center and the AV heading, respectively (see Fig.~\ref{fig:architecture}a), while they are in the actor frames in the two-stage models (see Fig.~\ref{fig:architecture}b-c). 
Then, $\mathcal{L}_{det}$ and $\mathcal{L}_{pred}$ can be optimized together in a joint training.

For single-stage models the detection and prediction values are both optimized in the first stage (Fig.~\ref{fig:architecture}a). 
On the other hand, when the first stage serves as a part of the two-stage architecture (Fig.~\ref{fig:architecture}b-c), $\mathcal{L}_{det}$ is optimized as a part of the first-stage output while $\mathcal{L}_{pred}$ is optimized in the second stage, discussed in the remainder of this section.





\subsection{Interaction using convolutions implicitly}
\label{method:icm}
In the previous section we discussed the first-stage feature extraction, that computes per-actor grid features which are then used as an input to the second-stage models to predict future motion. 
In this section we discuss how to compute the per-actor features better at capturing interactions:
\begin{itemize}
\item To capture relationship to nearby actors for the actor for whom the future trajectories are predicted (called the actor of interest), an input of the forecasting module can be a region covering the interacting actors and objects on the feature map, instead of only using the feature pixel. 
For the traffic use-case this interaction region (IR) should cover the area within which the objects should be paid attention to. 
Our results show that for vehicle actors, a large region ahead of the vehicle provides good context to model interaction.
\item To overcome rotational variance of convolutions, instead of cropping the IR features in the coordinate frame of the original BEV grid whose orientation is determined by AV, we define IR in the frame of the actor of interest (i.e., actor frame), in which the output trajectories are also defined (commonly referred to as RROI \cite{ma2017rroi}). 
Our results confirm the importance of rotational invariance in modeling interactions.
\item To effectively propagate non-local information of the interacting actors to the actor of interest, we can use an interactive CNN (ICNN) consisting of a few downsampling convolutional layers that eventually condense an IR comprising the actor of interest itself, its surrounding actors, and the environment, into a feature vector used as the final feature for this actor. 
\end{itemize}

As mentioned, the actor-centric feature map and the RROI techniques have been utilized in a number of applications~\cite{dp2018, casas2019spatially, djuric2020multixnet}, where it was found to lower displacement errors in trajectory forecasting tasks. 
In this paper we demonstrate that, by combining these ideas, convolutions are effective in modeling spatial interactions as well. 
Moreover, as shown by our experiments, by varying the parameters of these ingredients one can control the level of interaction modeling, providing further evidence that spatial interactions can be effectively captured by convolutions. 

The implementation of these three components are illustrated in the dashed box in Fig.~\ref{fig:architecture}b, which we refer to as the interaction convolutional module (ICM). 
For each actor we define a square IR around it, which is then used to crop actor-centric features from the global feature map using bilinear interpolation.
We vary the size, orientation, and the position of the actor in the IR to study their effects on the performance of interaction modeling (e.g., in the extreme case where the IR has no area, the cropped feature is just the feature pixel on the feature map). 
We choose a square IR to simplify the discussion. The length of the square side is referred to as {\it IR size} in the following discussion. 
Similarly, the ICNN module always consists of six ConvBs and one ResB to gradually reduce the cropped feature map to a 1D feature vector $f_c$ (e.g., if the crop size is $32 \times 32$, setting the strides of the last five ConvBs to 2 yields a $1$D vector; see Appendix for detailed discussion on crop sizes and ICNN design). 
The final multimodal classification and future trajectory regression in the actor frame are obtained from this $1$D vector via a single FC layer, one for each task.



\subsection{Interaction using graphs explicitly}
The purely convolutional approach described in the previous section provides implicit interaction modeling. 
To explicitly account for interactions, a common approach is the use of GNNs, discussed in this section.
As there exist many variants, we choose one of the more general approaches, the message passing neural network \cite{gilmer2017neural, wang2018non}, which has also been adapted to the motion forecasting problem~\cite{casas2019spatially}. 

Indicated by the dashed box in Fig.~\ref{fig:architecture}c, a fully connected graph comprises all of the $V$ actors (represented as nodes), with bi-directional edges between every two actors. 
The feature attribute $n_i$ of the $i$-th node is initialized by 
\begin{equation}
    n_{i}^0 = \mbox{MLP}_{init}(f_{ci}),
\end{equation} 
where $f_{ci}$ is the final feature vector of the $i$-th actor computed in the previous section. 
All multi-layer perceptrons (MLPs) in this GNN have two layers. The message passing at the $k$-th iteration via  edge from node $j$ to $i$ is given by
\begin{equation}
    m_{j\rightarrow i}^k = \mbox{MLP}_e^k([n_i^k, ~\mbox{rel}_{j\rightarrow i}, ~n_j^k, ~\mbox{rel}_{i\rightarrow j}]), 
\end{equation}
where $[\cdot]$ denotes concatenation. 
Unlike the implicit convolutional approach in the previous section where the relative spatial relation of actors are intrinsically represented within the crop, spatial relationships are additionally required in a graph representation.  
The relative geometric feature rel$_{j\rightarrow i}$ consisting of the coordinates and heading of actor $j$ in the frame of actor $i$, is computed as
\begin{equation}
    \mbox{rel}_{j\rightarrow i} = \mbox{MLP}_{\mbox{rel}}([x_{j\rightarrow i}, ~y_{j\rightarrow i}, ~\cos\theta_{j\rightarrow i}, ~\sin\theta_{j\rightarrow i}]).
\end{equation}
All of the messages sent to the $i$-th graph node are aggregated by a max-pooling operation, denoted as
\begin{equation}
\label{eq:pool}
    m_{i}^k = \mbox{Pool}_j (m_{j\rightarrow i}^k). 
\end{equation}
Finally, the node attribute is updated with a Gated Recurrent Unit (GRU) \cite{casas2019spatially,gilmer2017neural, wang2018non} whose hidden state is $n_i^k$ and the input is $m_{i}^k$, 
\begin{equation}
    n_{i}^{k+1} = \mbox{GRU} (n_{i}^{k}, m_{i}^k). 
\end{equation}
In general, the update iterates for $K$ times. 
Finally, multimodal classification and future trajectories for the actor are computed from $n_{i}^{k+1}$, as discussed in Section \ref{method:icm}.  
\begin{figure}[t]
    \centering%
    \includegraphics[width=0.5\linewidth]{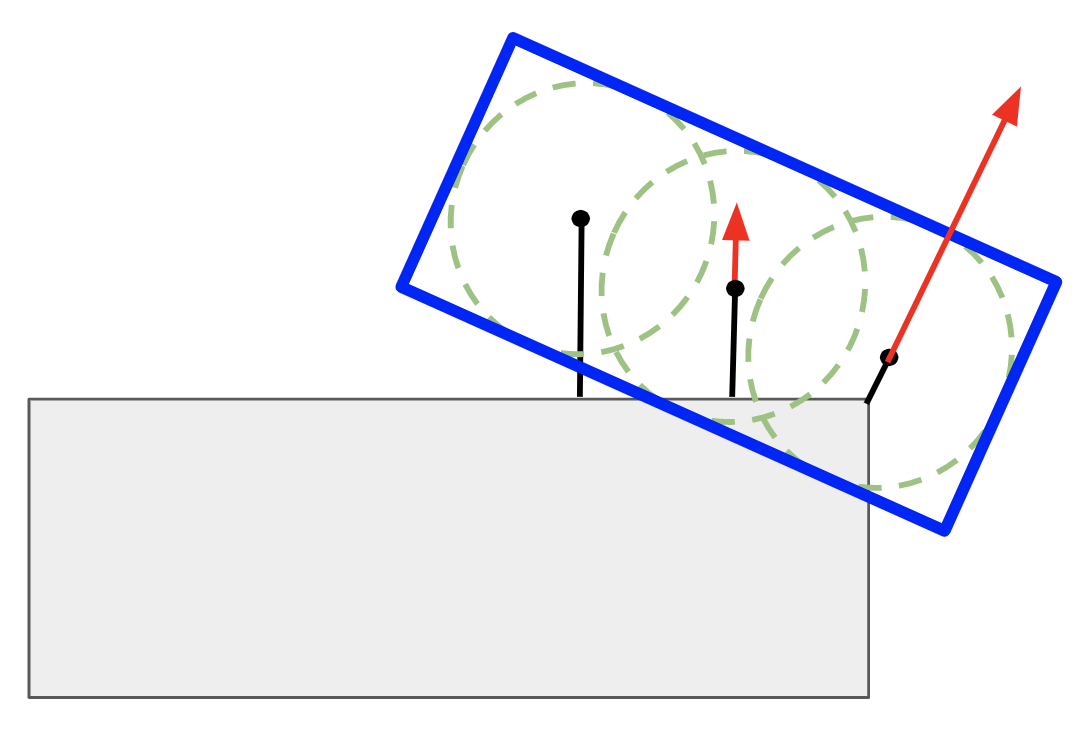}
    \caption{Schematic of interaction loss. The actor (blue) is approximated with 3 costing circles (green), with minimal distances (black) to an obstacle (grey) and resulting gradients (red).} 
    \label{fig:collision_loss}
\end{figure}

\subsection{Interaction loss}
In this section we introduce a novel interaction loss to improve interaction awareness of the model, which directly penalizes predicted forecast of an actor that overlap with static traffic objects (defined as objects with speed less than $0.2$m/s). 
Traffic objects comprise objects that a vehicle should avoid, including vehicles, cyclists, pedestrians, construction fences, etc.
At each prediction horizon, the predicted actor is approximated with $3$ inscribed costing circles, as illustrated in Fig.~\ref{fig:collision_loss}. 
The loss is then computed as
\begin{equation}
    \mathcal{L}_{col} = \frac{1}{3VH}\sum_{v}^V\sum_{n}^{N}\sum_{t=1}^H\sum_{l=1}^{L=3} \max (0, ~R_{vl} - d_{vntl}),
    \label{eq:interaction_loss}
\end{equation}
where $V$, $N$, $H$, and $L$ are the numbers of actors, non-moving obstacles, prediction time horizons, and costing circles, respectively. 
$R_{vl}$ is a radius of a costing circle (determined by the size of a ground-truth bounding box), while $d_{vntl}$ is a signed minimum distance between the $l$-th costing circle center of the $v$-th actor and the $n$-th obstacle bounding box at time $t$. 
The distance is negative when the center is inside the obstacle's bounding box. 

Note that the loss only considers overlaps between predicted trajectories and the ground-truth bounding boxes of static obstacles. 
Moving actors may have multimodal trajectory distributions, and it can be unclear when an overlap between the trajectories of two moving actors should be penalized by the loss.
In summary, when the costing circles overlap with an obstacle bounding box, the interaction loss would only back-propagate gradients through the predicted centroid and heading. 
The loss is added to the prediction loss $\mathcal{L}_{pred}$ where it is applied to the $\hat m$-th predicted trajectory, and optimized jointly in the end-to-end training. 

\section{EXPERIMENTS}
\label{sect:experiments}
\begin{figure*}[!htbp]
    \centering%
    \includegraphics[width=0.9\textwidth]{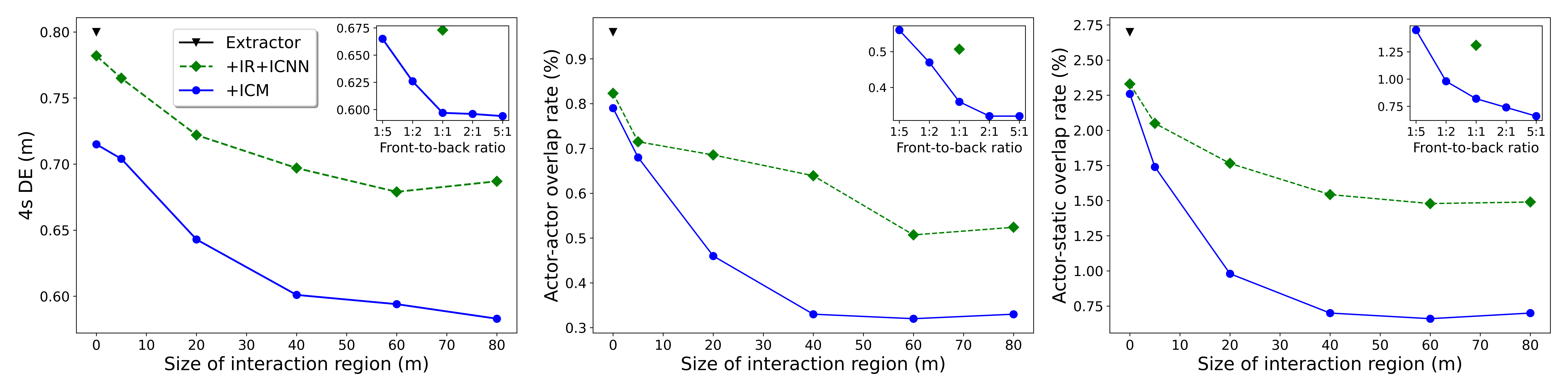}
    \caption{Effects of the ICM components: interaction region (IR), interaction CNN (ICNN), and actor frame (AF). {\bf Extractor} is the single-stage model; 
    {\bf +IR+ICNN} represents the two-stage model that defines IRs in the AV frame; {\bf +ICM} (i.e., +IR+ICNN+AF) is the proposed two-stage model that defines IRs in the actor frame. All IRs have a fixed 5:1 front-to-back ratio, unless specified otherwise. Inset: models with a fixed IR size of 60m and varied front-to-back ratios.}
    \label{fig:ICM}
\end{figure*}

{\bf Input and output.} The considered area is of size $150 \times 100 \times 3.2$m, centered at AV and discretized as a $960 \times 640 \times 16$ grid into which the LiDAR sweep information is encoded. 
The input contains $10$ LiDAR sweeps collected at 0.1s interval, as well as a semantic HD map from the current timestamp. 
The models detect the vehicle actors at the current time step and forecast their trajectories at future time horizons $t \in \{0.1, 0.2, \ldots, 4.0s\}$. 
Non-maximum suppression (NMS)~\cite{neubeck2006efficient} with Intersection over Union (IoU) threshold set at 0.1 is applied in order to eliminate duplicate detections.

{\bf Metrics.} 
The studies are focused on prediction accuracy and interaction performance. 
IoU threshold for object detection matching is set at $0.5$. 
We observe that the detection performance changes little in all of the considered models reported in the paper, with average precision at $94.0 \pm 0.4$. 
Furthermore, we ensure equal numbers of trajectories are considered in the metrics by adjusting the detection probability threshold at a fixed recall of $0.8$ \cite{casas2019spatially}. 
Each actor has 3 predicted trajectory modes, and we assign the trajectory of the most probable mode to the actor in the following metric computation. 
We use displacement error (DE) at $4$s to measure the prediction accuracy, averaged over all actors. 

To quantify the interaction performance of the models we consider two overlap metrics in our experiments (additional results of other metrics are provided in Appendix): 
\begin{itemize}
\item Actor-actor overlap rate is the percentage of predicted trajectories of detected actors overlapping with predicted trajectories of other detected actors. 
\item Actor-static overlap rate is the percentage of predicted trajectories of detected actors overlapping with ground-truth static traffic objects. 
\end{itemize}
An actor overlap is defined as an intersection-over-obstacle-polygon of more than $0.05$ at any point of the 4s-long trajectory, set to this value to eliminate false positive overlaps due to small noise in the labeled bounding boxes.

{\bf Data.}
\label{main:data}
We conducted an evaluation on a large in-house data set, 
containing $19{,}000$ scenes of $25$s each and collected across several cities in North America with high-quality $10$Hz annotations. 
To mitigate the metric variance of the sparse overlaps, (1) a large split of $5{,}000$ scenes is left out for testing; (2) the test frames in scenes have a temporal spacing of $2$s to avoid counting the same overlaps multiple times; and (3) the training and test sets are split geographically to prevent models from memorizing the same static obstacles and environment. 
Using this larger data set, as opposed to using popular open-sourced data sets that are significantly smaller, allows for lower metric variance and deriving more general conclusions. 
Lastly, by using the same input, backbone network, loss functions, and training settings, our studies contrast the interaction modeling approaches. This allows us to focus on the relative performance of these approaches, as opposed to comparing independent models where ensuring similar network capacities and equally well-tuned hyper-parameters are typically challenging tasks. 
In Appendix we provide details on the data sets, metric variance, and comparison to other motion forecasting models on public data sets.


\subsection{Results}
\label{sect:results_quant}
{\bf Interaction using convolutions.} 
The performance of the single-stage model (Fig.~\ref{fig:architecture}a) that contains only the feature extractor is shown in Fig.~\ref{fig:ICM} ({\bf Extractor}, black). 
The {\bf +IR+ICNN} (green) curve shows the performance of the two-stage model without rotating the interaction region for each actor into the actor frame. 
In particular, starting from the 1D per-actor feature map vectors ($0$m), we increase the IR size to $80$m.  
By cropping larger feature map regions that contain
more interacting actors and surrounding context, displacement error and forecasted overlap rates decrease. 

\begin{figure*}[!htbp]
    \centering
    \includegraphics[width=0.90\textwidth]{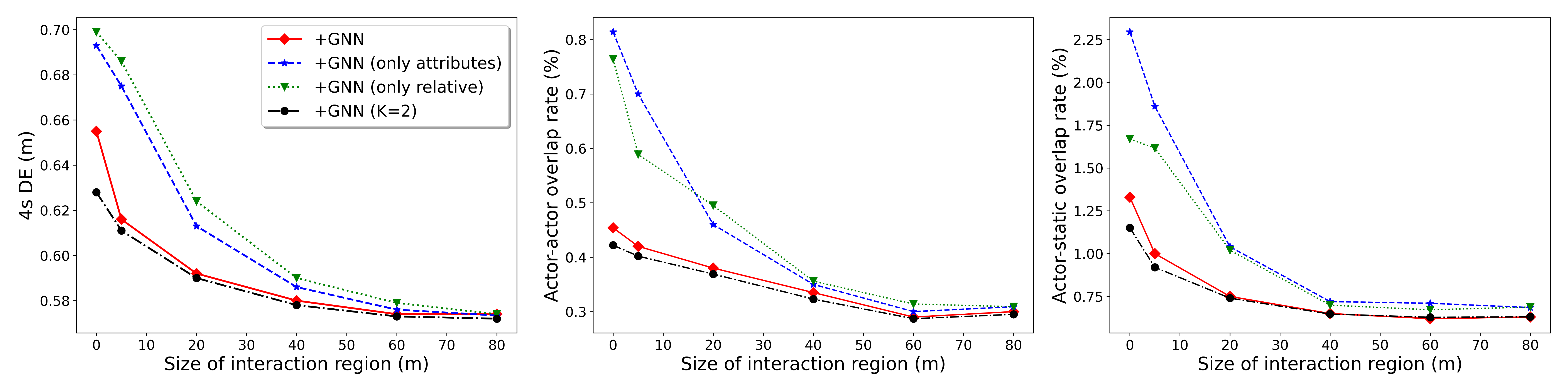}
    \caption{Performance from adding a GNN on top of the ICM (Fig.~\ref{fig:architecture}c) for different ICM interaction region sizes. {\bf +GNN (only attributes)} encodes only node attributes in the graph edges; {\bf +GNN (only relative)} encodes only relative locations and orientations in the graph edges.}
    \label{fig:GNN}
\end{figure*}

\begin{figure*}[!htbp]
    \centering
    \includegraphics[width=0.90\textwidth]{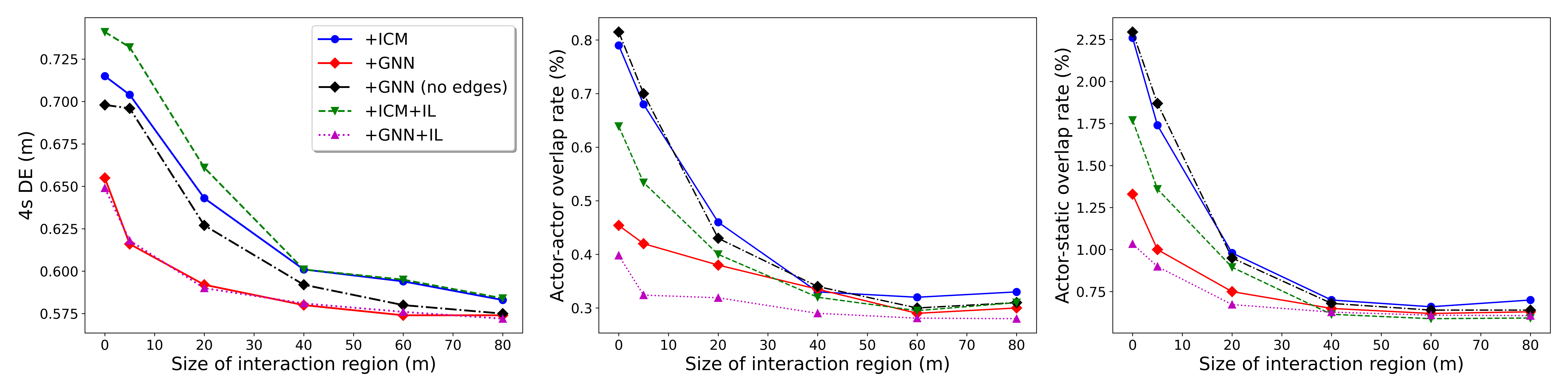}
    \caption{Comparison of ICM and GNN (which includes ICM). {\bf +GNN (no edges)} is identical to {\bf +GNN} except the graph edges are cut off. {\bf +IL} represents models trained with the additional interaction loss. Note that comparing {\bf +ICM} at  large  IR  (interaction  is  modeled  by  ICM)  against {\bf +GNN} at  small  IR  (interaction  is  modeled  by  GNN) shows that a pure ICM can outperform a pure GNN in modeling interactions.}
    \label{fig:compare}
\end{figure*}

We then rotate IRs to match estimated actor orientation instead of using the common AV frame ({\bf +ICM}, blue). 
For zero IR size (i.e., a cropped feature is still the feature pixel), we observe DE drops significantly compared to the model using the AV frame with zero IR size (green). 
This has been explained previously as a benefit of a standardized output representation \cite{djuric2020multixnet}. 
Although defining the IR in the actor frame reduces rotational variance, the zero-size IR covers no interacting actors and we thus observe little change in the actor overlap rates. Here a lower DE is not less correlated to better interaction modeling. As the IR is increased in size, both DE and the interaction metrics improve dramatically. 
Crop sizes of beyond $60$m show no further improvement, likely because the majority of interacting actors and obstacles are already included within the  $60$m region.

In all of the IRs above we have fixed the front-to-back ratio to 5:1, meaning an IR of size $60$m includes $50$m ahead and $10$m behind the actor. 
In Fig.~\ref{fig:ICM} inset we fix the total size at $60$m, and vary the front-to-back ratio (blue). 
As the vast majority of actors are moving forward, we can see that placing more of the IR ahead of the actor improves interaction modeling. 
It is interesting to note the divergence between DE and overlap rates again: after the front-to-back ratio is above 1:1, the overlap rates continue to drop marginally, while the DE improvement stops.
Even for the actor-centered IR (inset, green), not rotating the IR to match the actor orientation yields worse DE and overlap rates, which further confirms the importance of removing rotational variance for interaction modeling using convolutions.
From Fig. \ref{fig:ICM}, we observe that by cropping an actor-frame defined region of the feature map and then applying convolutions improves forecasting and interaction modeling considerably. 
Strong dependence of overlap rates on IR size provides evidence that convolutions are effectively capturing interactions once other actors are inside the IR.  

{\bf Interaction using graphs.} 
As illustrated in Fig. \ref{fig:architecture}c, for these experiments we add a GNN after the ICM. 
Note that, as discussed earlier, setting the IR size to $0$m deactivates the ICM while retaining the benefit of reduced rotational variance. 
For zero IR size (Fig. \ref{fig:GNN}, {\bf +GNN}, red) we see that the GNN indeed improves DE and overlap rates significantly as compared to the models without designated interaction modeling capability in Fig. \ref{fig:ICM} ({\bf +ICM}, $0$m). 
Notably, even when GNN is utilized, we observe that ICM can still provide additional performance improvements as we gradually increase ICM's interaction modeling by expanding the IR size.   
We also examine the benefit of the hand-crafted relative geometries in the graph edges. 
When the IR is small (i.e., ICM is limited), keeping only the node attributes $n_i$ (blue) or relative geometries $\mbox{rel}_{i, j}$ (green) significantly damages the graph modeling. 
For large IR sizes, the difference between the three graph models becomes minor, suggesting that with larger feature crops the ICM has effectively compensated for missing GNN features.

The GNNs in the models above are single-iteration. 
We also evaluated the effect of increasing the GNN iterations to $K=2$. 
An additional iteration (i.e., $K=2$) reduces DE and overlap rates further by a small amount when the IR size is small, which could be explained by the well-known bottleneck phenomenon of GNNs~\cite{alon2021bottleneck} and the fact that the graph is fully connected. 
This improvement is negligible for all but the smallest IRs, and no further exploration of additional iterations is provided below.

{\bf Convolutions vs. graphs for interaction.} 
In Fig. \ref{fig:compare} we compare the implicit ICM (blue) and explicit GNN (red) approaches. 
With zero IR size (where ICM is effectively off) the gain of adding GNN is significant. 
However, as the IR grows, we observe that the performance gap steadily narrows. 
In other words, while turning on ICM (by increasing IR size) can further improve the performance of GNN models, adding a graph to an ICM with a sufficiently large IR provides only minor benefits.
To understand the gaps between {\bf +ICM} and {\bf +GNN} with large IR sizes, we study a graph-less model ({\bf +GNN (no edge)}, black) created by removing graph edges in {\bf +GNN}. 
For large IRs, the graph-less model matches the performance of {\bf +GNN}, suggesting that explicit interaction graph of GNN contributes little to the performance.
Thus, the gaps between {\bf +ICM} and {\bf +GNN} for larger IR sizes are mainly due to extra network capacity of GNN. 
Lastly, we see that comparing {\bf +ICM} at large IR (i.e., interaction modeled by ICM) against {\bf +GNN} at small IR (i.e., interaction modeled by GNN) shows that a pure ICM can outperform a pure GNN in modeling interactions.

{\bf Interaction loss.} 
We can also see that adding the interaction loss (Eq. \ref{eq:interaction_loss}) reduces the overlap between actors' predicted trajectories for both interaction modeling approaches (green and magenta in Fig. \ref{fig:compare}). 
The improvement is significant for smaller IRs, which may be due to the fact that the smaller IRs do not provide enough information to model the interactions effectively, benefiting more from this added supervision. 
On the other hand, when interaction is modeled more effectively through larger IR, the loss is sparser and thus contributes less. 
Interestingly, the interaction loss does not affect DE results except for ICM models at small IR where the interaction modeling is limited. 

\begin{figure}[!t]
    \centering
    \includegraphics[width=1.0\linewidth]{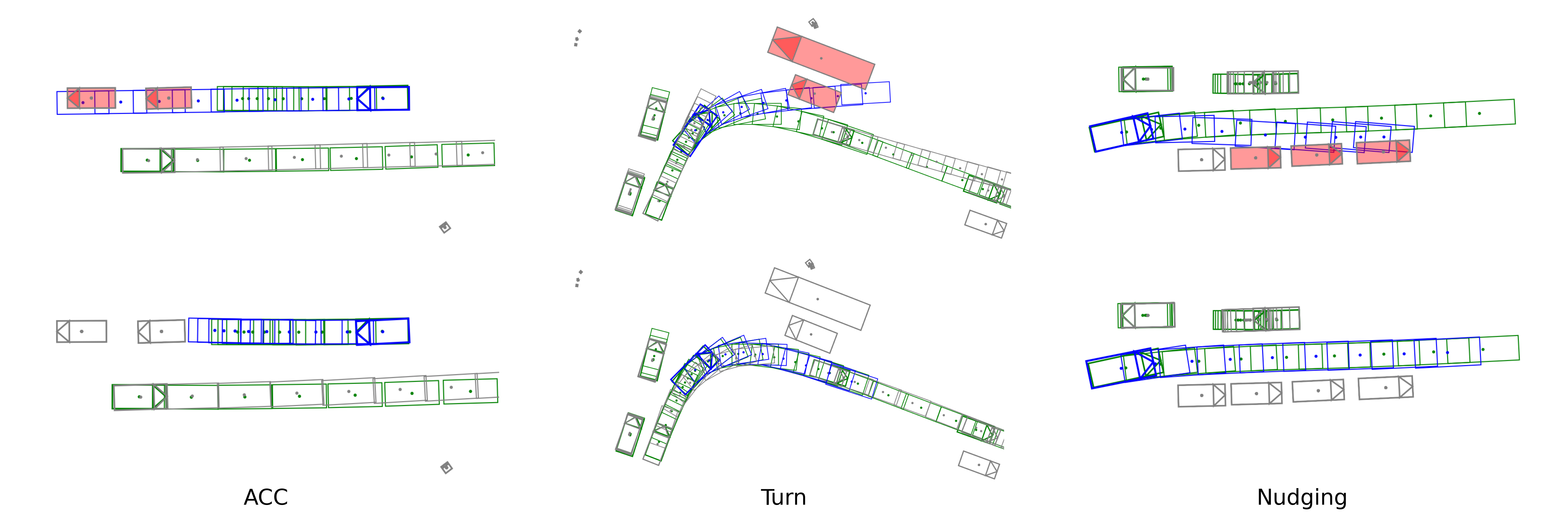}
    \caption{Predicted trajectories sampled at 2Hz of baseline (top) and ICM (bottom). red: overlapped obstacles; blue: forecasts of the actors of interest; grey: forecasts of other actors; green: labels (also see attached videos in the Appendix).}
    \label{fig:obstacle}
\end{figure}

{\bf Maneuver-specific qualitative results.} 
In Fig. \ref{fig:obstacle} we present a comparison of the baseline ICM model with $0$m size (that has no designated interaction modeling) and the ICM model with $60$m size on three typical maneuvers observed in interacting scenarios: adaptive cruise control (ACC), turn, and nudging.
We note that the $0$m model in all cases incorrectly predicts overlapping trajectories.
In the ACC case the ICM model correctly predicts that the vehicle would decelerate and queue after others, while in the turn case it outputs a trajectory that follows the lane and avoids overlapping with the vehicles after the turn.
In the nudging case the vehicle motion starts with considerable curvature, the forecast correctly reduces the curvature and straightens the trajectory to avoid the parked cars.
We also examined the results of adding GNNs on +ICM ($60$m) on these maneuvers, and observed no significant difference. 


{\bf Inference time.} 
The baseline model that includes the feature extractor and other parts such as input pre-processing and output post-processing takes $45.6$ms per frame. 
Next we measure the additional time costs of adding the ICM and GNN modules to the baseline model, shown in Table \ref{tab:latency}. 
The ICM of zero IR size adds an additional $5.2$ms, which includes processing of the feature pixel and computation of the final output.
ICM with non-zero size uses convolutions and bilinear feature cropping, operations that have been optimized in current GPU software and hardware. 
As a result, even the largest $80$m ICM is only a few milliseconds slower than the $0$m ICM.
Lastly, the GNN itself takes $46.9$ms, multiple times slower than the slowest ICM.  
This is consistent with earlier results showing GNN inference may be inefficient resulting in higher latency~\cite{kiningham2020grip}.
Coupled with the earlier results showing that modeling interaction using convolutions can give competitive performance compared to GNNs, we see that the convolutional approach represents an efficient and practical alternative to GNNs.

\begin{table} [t!]
\small 
\caption{Inference times of modules (tested on Nvidia Titan RTX)
}
\label{tab:latency}
\centering
{
  \begin{tabular}{lccccccccc}
    {\bf Module} & {\bf IR size [m]} & {\bf Inference [ms]}\\
    \hline
    \rowcolor{lightgray}
    ICM & 0 & 5.2 \\
    ICM & 80 & 8.1 \\
    \rowcolor{lightgray}
    GNN & - & 46.9 \\  
    \hline
\end{tabular}
}
\end{table}
\section{LIMITATIONS AND SOCIAL IMPACT}
\label{sect:limitations}

While graphs represent a general approach to modeling various relations, this work shows that convolutions are also effective when it comes to modeling spatial interaction. 
However, in Euclidean space where CNNs are applied, some information such as driver-driver vocal interactions are not currently represented. Besides, as our studies are limited to the 2D AV application, it is unanswered whether convolutions are still effective in capturing interactions in 3D space (such as for modeling human body motion). 
Additionally, the overlap rates over trajectories enable us to explicitly evaluate quality of the interaction modeling. However, the metric sparsity leads to requirement for having larger test data to ensure the obtained results are meaningful, which limits a wider use of the metric. 
Lastly, neither of the considered approaches guarantees zero overlaps over the trajectories, and CNN models also often have lower interpretability than the GNN approaches.


\section{CONCLUSION}
\label{sect:conclusion}

We compared and contrasted 2D convolutional and graph neural networks for the task of spatial interaction modeling, providing empirical evidence that under certain conditions convolutional networks can reach comparable performance to the state-of-the-art GNNs (e.g., by modifying IR), thus allowing similar motion forecasting accuracy and interaction modeling while maintaining reduced latency and model complexity of the model.
We analyzed common components of the interaction approaches, leading to a better understanding of how each benefits the interaction modeling.
Moreover, we introduced a novel interaction-aware loss and showed its impact on the considered approaches.
Our work presents a basis for wider use of convolutional layers for the task of spatial interaction, providing evidence that the gap between convolutional models and more complex and computationally expensive GNN models may not be as large as previously suspected.


\balance
{\small
\bibliographystyle{ieee_fullname}
\bibliography{egbib}

\begin{thebibliography}{10}\itemsep=-1pt

\bibitem{alahi2016social}
Alexandre Alahi, Kratarth Goel, Vignesh Ramanathan, Alexandre Robicquet, Li
  Fei-Fei, and Silvio Savarese.
\newblock Social lstm: Human trajectory prediction in crowded spaces.
\newblock In {\em Proceedings of the IEEE conference on computer vision and
  pattern recognition}, pages 961--971, 2016.

\bibitem{alon2021bottleneck}
Uri Alon and Eran Yahav.
\newblock On the bottleneck of graph neural networks and its practical
  implications.
\newblock {\em arXiv preprint arXiv:2006.05205}, 2021.

\bibitem{battaglia2016interaction}
Peter~W Battaglia, Razvan Pascanu, Matthew Lai, Danilo Rezende, and Koray
  Kavukcuoglu.
\newblock Interaction networks for learning about objects, relations and
  physics.
\newblock {\em arXiv preprint arXiv:1612.00222}, 2016.

\bibitem{caesar2020nuscenes}
Holger Caesar, Varun Bankiti, Alex~H Lang, Sourabh Vora, Venice~Erin Liong,
  Qiang Xu, Anush Krishnan, Yu Pan, Giancarlo Baldan, and Oscar Beijbom.
\newblock nuscenes: A multimodal dataset for autonomous driving.
\newblock In {\em Proceedings of the IEEE/CVF Conference on Computer Vision and
  Pattern Recognition}, pages 11621--11631, 2020.

\bibitem{casas2019spatially}
Sergio Casas, Cole Gulino, Renjie Liao, and Raquel Urtasun.
\newblock Spatially-aware graph neural networks for relational behavior
  forecasting from sensor data.
\newblock {\em arXiv preprint arXiv:1910.08233}, 2019.

\bibitem{casas2018intentnet}
Sergio Casas, Wenjie Luo, and Raquel Urtasun.
\newblock Intentnet: Learning to predict intention from raw sensor data.
\newblock In {\em Conference on Robot Learning}, pages 947--956, 2018.

\bibitem{cui2019multimodal}
Henggang Cui, Vladan Radosavljevic, Fang-Chieh Chou, Tsung-Han Lin, Thi Nguyen,
  Tzu-Kuo Huang, Jeff Schneider, and Nemanja Djuric.
\newblock Multimodal trajectory predictions for autonomous driving using deep
  convolutional networks.
\newblock In {\em 2019 International Conference on Robotics and Automation
  (ICRA)}, pages 2090--2096. IEEE, 2019.

\bibitem{Deo2018Conv}
Nachiket Deo and Mohan~M. Trivedi.
\newblock Convolutional social pooling for vehicle trajectory prediction.
\newblock {\em CoRR}, abs/1805.06771, 2018.

\bibitem{Diehl2021graph}
Frederik Diehl, Thomas Brunner, Michael Truong{-}Le, and Alois~C. Knoll.
\newblock Graph neural networks for modelling traffic participant interaction.
\newblock {\em CoRR}, abs/1903.01254, 2019.

\bibitem{djuric2020multixnet}
Nemanja Djuric, Henggang Cui, Zhaoen Su, Shangxuan Wu, Huahua Wang, Fang-Chieh
  Chou, Luisa~San Martin, Song Feng, Rui Hu, Yang Xu, et~al.
\newblock Multinet: Multiclass multistage multimodal motion prediction.
\newblock In {\em Proceedings of the IEEE Intelligent Vehicles Symposium (IV)},
  2020.

\bibitem{dp2018}
Nemanja Djuric, Vladan Radosavljevic, Henggang Cui, Thi Nguyen, Fang-Chieh
  Chou, Tsung-Han Lin, and Jeff Schneider.
\newblock Short-term motion prediction of traffic actors for autonomous driving
  using deep convolutional networks.
\newblock {\em arXiv preprint arXiv:1808.05819}, 2018.

\bibitem{engelmann2020dilated}
Francis Engelmann, Theodora Kontogianni, and Bastian Leibe.
\newblock Dilated point convolutions: On the receptive field size of point
  convolutions on 3d point clouds.
\newblock In {\em 2020 IEEE International Conference on Robotics and Automation
  (ICRA)}, pages 9463--9469. IEEE, 2020.

\bibitem{fout2017protein}
Alex Fout, Jonathon Byrd, Basir Shariat, and Asa Ben-Hur.
\newblock Protein interface prediction using graph convolutional networks.
\newblock In {\em Advances in neural information processing systems}, pages
  6530--6539, 2017.

\bibitem{gao2020vectornet}
Jiyang Gao, Chen Sun, Hang Zhao, Yi Shen, Dragomir Anguelov, Congcong Li, and
  Cordelia Schmid.
\newblock Vectornet: Encoding hd maps and agent dynamics from vectorized
  representation.
\newblock {\em arXiv preprint arXiv:2005.04259}, 2020.

\bibitem{gilmer2017neural}
Justin Gilmer, Samuel~S Schoenholz, Patrick~F Riley, Oriol Vinyals, and
  George~E Dahl.
\newblock Neural message passing for quantum chemistry.
\newblock {\em arXiv preprint arXiv:1704.01212}, 2017.

\bibitem{gupta2018social-gan}
Agrim Gupta, Justin Johnson, Li Fei{-}Fei, Silvio Savarese, and Alexandre
  Alahi.
\newblock Social {GAN:} socially acceptable trajectories with generative
  adversarial networks.
\newblock {\em CoRR}, abs/1803.10892, 2018.

\bibitem{hamaguchi2017knowledge}
Takuo Hamaguchi, Hidekazu Oiwa, Masashi Shimbo, and Yuji Matsumoto.
\newblock Knowledge transfer for out-of-knowledge-base entities: A graph neural
  network approach.
\newblock {\em arXiv preprint arXiv:1706.05674}, 2017.

\bibitem{hamilton2017inductive}
Will Hamilton, Zhitao Ying, and Jure Leskovec.
\newblock Inductive representation learning on large graphs.
\newblock In {\em Advances in neural information processing systems}, pages
  1024--1034, 2017.

\bibitem{he2016identity}
Kaiming He, Xiangyu Zhang, Shaoqing Ren, and Jian Sun.
\newblock Identity mappings in deep residual networks.
\newblock In {\em European conference on computer vision}, pages 630--645.
  Springer, 2016.

\bibitem{boris2018Modeling}
Boris Ivanovic and Marco Pavone.
\newblock Modeling multimodal dynamic spatiotemporal graphs.
\newblock {\em CoRR}, abs/1810.05993, 2018.

\bibitem{khalil2017learning}
Elias Khalil, Hanjun Dai, Yuyu Zhang, Bistra Dilkina, and Le Song.
\newblock Learning combinatorial optimization algorithms over graphs.
\newblock In {\em Advances in neural information processing systems}, pages
  6348--6358, 2017.

\bibitem{kingma2014adam}
Diederik~P Kingma and Jimmy Ba.
\newblock Adam: A method for stochastic optimization.
\newblock {\em arXiv preprint arXiv:1412.6980}, 2014.

\bibitem{kiningham2020grip}
Kevin Kiningham, Christopher Re, and Philip Levis.
\newblock Grip: A graph neural network accelerator architecture.
\newblock {\em arXiv preprint arXiv:2007.13828}, 2020.

\bibitem{kipf2018neural}
Thomas Kipf, Ethan Fetaya, Kuan-Chieh Wang, Max Welling, and Richard Zemel.
\newblock Neural relational inference for interacting systems.
\newblock In {\em International Conference on Machine Learning}, pages
  2688--2697. PMLR, 2018.

\bibitem{kipf2016semi}
Thomas~N Kipf and Max Welling.
\newblock Semi-supervised classification with graph convolutional networks.
\newblock {\em arXiv preprint arXiv:1609.02907}, 2016.

\bibitem{lin2017focalloss}
T.-Y. Lin, P. Goyal, R. Girshick, K. He, and P. Dollar.
\newblock Focal loss for dense object detection.
\newblock In {\em ICCV}, 2017.

\bibitem{luo2018fast}
Wenjie Luo, Bin Yang, and Raquel Urtasun.
\newblock Fast and furious: Real time end-to-end 3d detection, tracking and
  motion forecasting with a single convolutional net.
\newblock In {\em Proc. of the IEEE CVPR}, pages 3569--3577, 2018.

\bibitem{lowe2020amortized}
Sindy Löwe, David Madras, Richard Zemel, and Max Welling.
\newblock Amortized causal discovery: Learning to infer causal graphs from
  time-series data.
\newblock {\em arXiv preprint arXiv:2006.10833}, 2020.

\bibitem{ma2017rroi}
Jianqi Ma, Weiyuan Shao, Hao Ye, Li Wang, Hong Wang, Yingbin Zheng, and
  Xiangyang Xue.
\newblock Arbitrary-oriented scene text detection via rotation proposals.
\newblock {\em CoRR}, abs/1703.01086, 2017.

\bibitem{meirom2020stop}
Eli~A. Meirom, Haggai Maron, Shie Mannor, and Gal Chechik.
\newblock How to stop epidemics: Controlling graph dynamics with reinforcement
  learning and graph neural networks.
\newblock {\em arXiv preprint arXiv:2010.05313}, 2020.

\bibitem{neubeck2006efficient}
Alexander Neubeck and Luc Van~Gool.
\newblock Efficient non-maximum suppression.
\newblock In {\em 18th International Conference on Pattern Recognition
  (ICPR'06)}, volume~3, pages 850--855. IEEE, 2006.

\bibitem{NEURIPS2019_9015}
Adam Paszke, Sam Gross, et~al.
\newblock Pytorch: An imperative style, high-performance deep learning library.
\newblock In H. Wallach, H. Larochelle, A. Beygelzimer, F. d\textquotesingle
  Alch\'{e}-Buc, E. Fox, and R. Garnett, editors, {\em Advances in Neural
  Information Processing Systems 32}, pages 8024--8035. Curran Associates,
  Inc., 2019.

\bibitem{Qasim_2019}
Shah~Rukh Qasim, Jan Kieseler, Yutaro Iiyama, and Maurizio Pierini.
\newblock Learning representations of irregular particle-detector geometry with
  distance-weighted graph networks.
\newblock {\em The European Physical Journal C}, 79(7), Jul 2019.

\bibitem{Rhinehart2019PRECOG}
Nicholas Rhinehart, Rowan McAllister, Kris~M. Kitani, and Sergey Levine.
\newblock {PRECOG:} prediction conditioned on goals in visual multi-agent
  settings.
\newblock {\em CoRR}, abs/1905.01296, 2019.

\bibitem{sadeghian2017car-net}
Amir Sadeghian, Ferdinand Legros, Maxime Voisin, Ricky Vesel, Alexandre Alahi,
  and Silvio Savarese.
\newblock Car-net: Clairvoyant attentive recurrent network.
\newblock {\em CoRR}, abs/1711.10061, 2017.

\bibitem{sanchez2018graph}
Alvaro Sanchez-Gonzalez, Nicolas Heess, Jost~Tobias Springenberg, Josh Merel,
  Martin Riedmiller, Raia Hadsell, and Peter Battaglia.
\newblock Graph networks as learnable physics engines for inference and
  control.
\newblock {\em arXiv preprint arXiv:1806.01242}, 2018.

\bibitem{schlichtkrull2017modeling}
Michael Schlichtkrull, Thomas~N Kipf, Peter Bloem, Rianne Van Den~Berg, Ivan
  Titov, and Max Welling.
\newblock Modeling relational data with graph convolutional networks.
\newblock In {\em European semantic web conference}, pages 593--607. Springer,
  2018.

\bibitem{su2020temporallycontinuous}
Zhaoen Su, Chao Wang, Henggang Cui, Nemanja Djuric, Carlos Vallespi-Gonzalez,
  and David Bradley.
\newblock Temporally-continuous probabilistic prediction using polynomial
  trajectory parameterization.
\newblock In {\em IEEE International Conference on Robotics and Automation
  (IROS)}, 2021.

\bibitem{wang2018non}
Xiaolong Wang, Ross Girshick, Abhinav Gupta, and Kaiming He.
\newblock Non-local neural networks.
\newblock In {\em Proceedings of the IEEE conference on computer vision and
  pattern recognition}, pages 7794--7803, 2018.

\bibitem{xu2020inductive}
Da Xu, Chuanwei Ruan, Evren Korpeoglu, Sushant Kumar, and Kannan Achan.
\newblock Inductive representation learning on temporal graphs.
\newblock {\em arXiv preprint arXiv:2002.07962}, 2020.

\end{thebibliography}
}


\renewcommand{\appendixpagename}{\centering Convolutions for Spatial Interaction Modeling\\- Appendix -}
\renewcommand{\thefigure}{S\arabic{figure}}
\setcounter{figure}{0} 
\onecolumn

\begin{appendices}
    In the appendix, we provide complete details on the model and experiment implementation to facilitate reproducing the proposed approaches and models: 
the network design of the first stage backbone (Section \ref{sup:extractor}), 
the ICNN network designs (Section \ref{sup:icnn}), 
the GNN network design (Section \ref{sup:gnn}), 
the training setup (Section \ref{sup:setting}), 
the data set choices (\ref{sup:data}), and 
the metric design and metric variances (Section \ref{sup:metric}).

We also provide additional quantitative and qualitative results: 
the experimental results focused on forecasted actor trajectories intersecting with non-vehicle traffic objects (Section \ref{sup:non-veh}) and 
videos (Section \ref{sup:video}).


\section{The extractor backbone network}
\label{sup:extractor}
\begin{figure*}[ht!]
\vspace{0.3cm}
    \centering
    \includegraphics[width=0.5\textwidth]{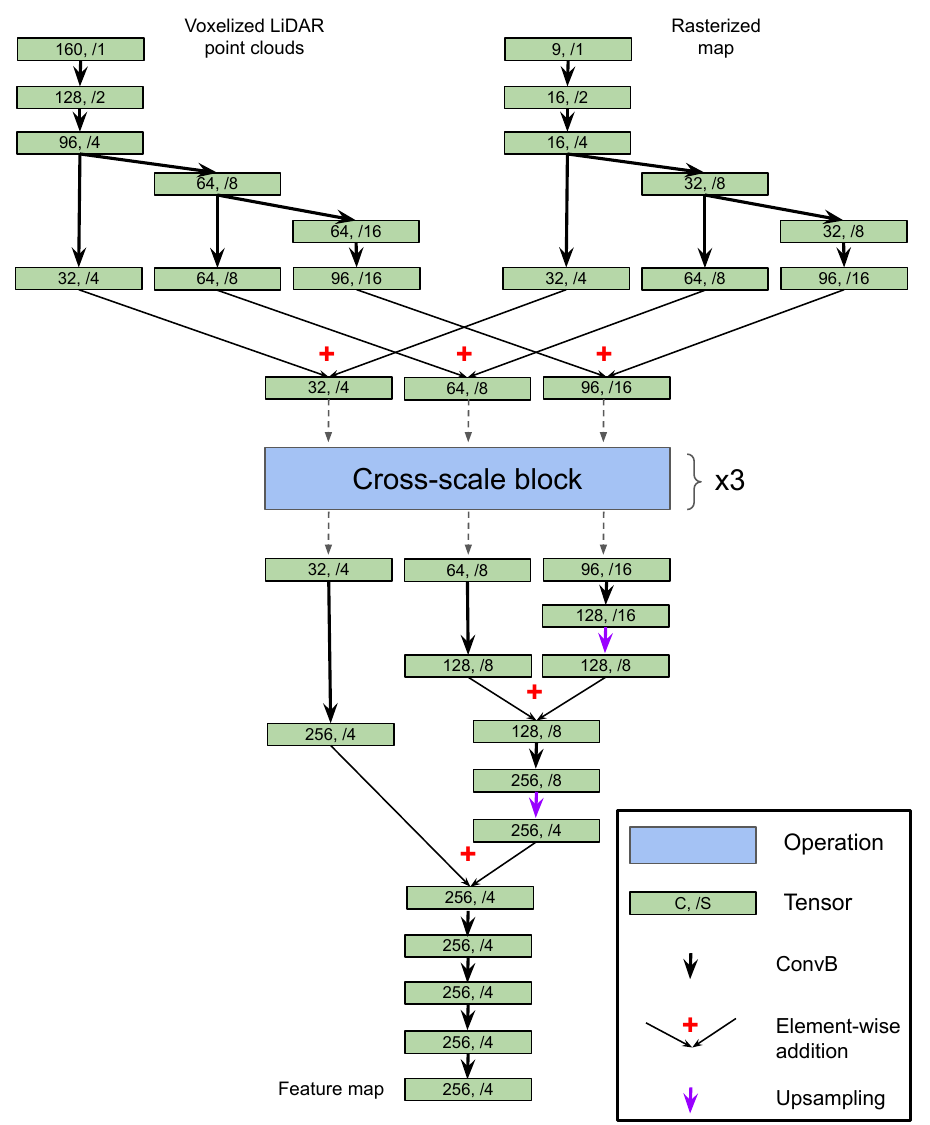}
    \caption{Multi-scale network design of the feature extractor. As illustrated in Fig.~\ref{fig:architecture}, the inputs are voxelized LiDAR point clouds and rasterized map, while the network output is a feature map of size $(256, W/4, L/4)$, where $W$ and $L$ are the grid width and length of the input BEV representation, respectively. The green boxes labeled as $C, /S$ represent tensors where $C$ and $S$ represent the number of channels and down-sampled scale relative to the input size, respectively. The operations connecting two tensors are {\bf ConvB}s, except for the specified up-sampling,  element-wise addition, and the cross-scale block. The cross-scale block, detailed in Fig. \ref{fig:cross_scale}, is repeated 3 times.}
    \label{fig:network}
\vspace{0.2cm}
\end{figure*}
\begin{figure*}[ht!]
    \centering
    \includegraphics[width=0.85\textwidth]{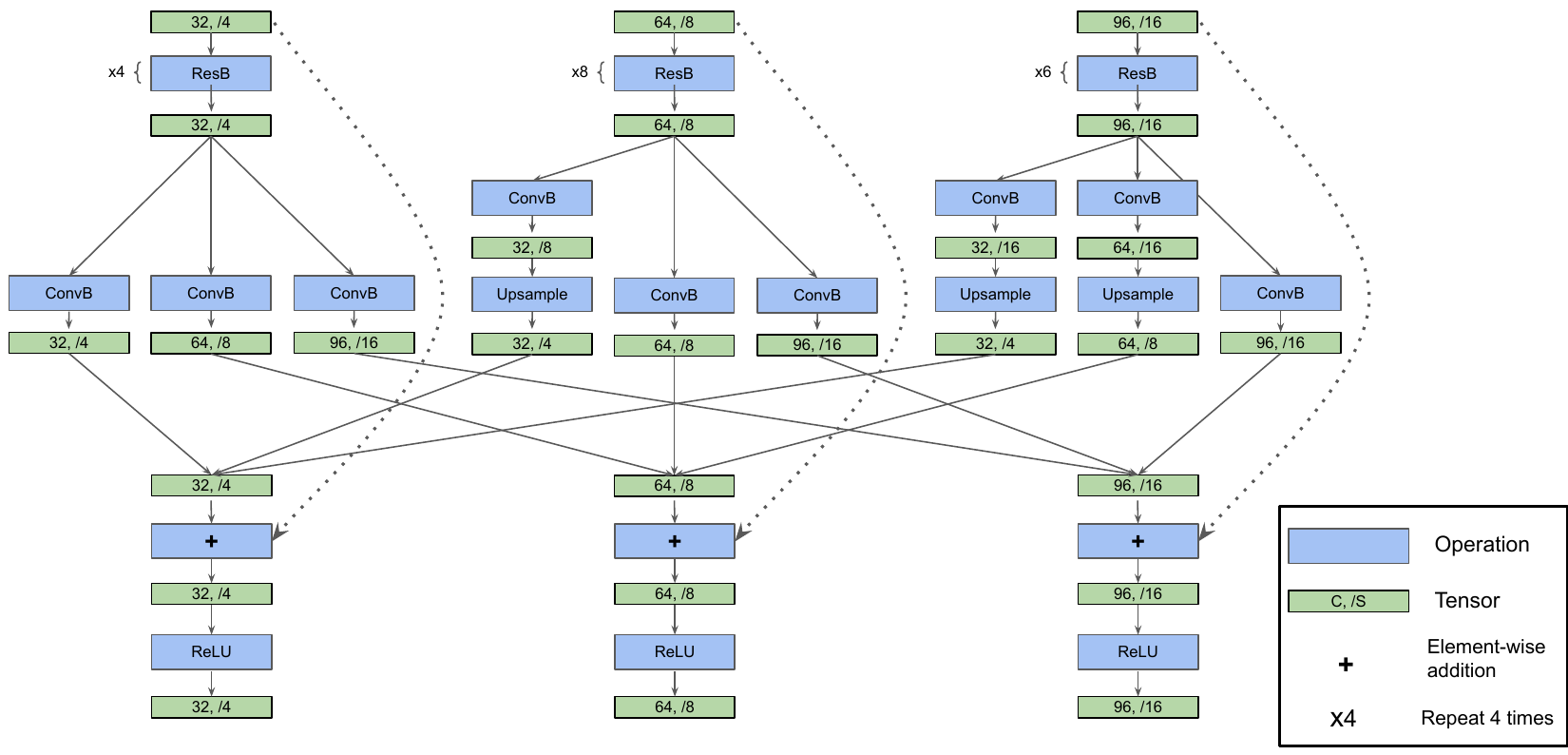}
    \caption{The cross-scale block}
    \label{fig:cross_scale}
\vspace{0.2cm}
\end{figure*}

In Fig. \ref{fig:network} we provide full and detailed design of the CNN feature extractor used in all of the models studied (see the high-level overview in Fig.~\ref{fig:architecture}). 
We note that the multi-scale design (as indicated by $/1$, $/2$, $/4$, $/8$, and $/16$, where the numbers represent the down-sampling scales relative to the input size) and cross-scale blocks (see Fig. \ref{fig:cross_scale}) already encourage a large receptive field. 
Nevertheless, the experimental results presented in the main paper show that such single-stage CNN architecture still models the spatial interaction less effectively. 
By adding either the shallow ICNN or the GNN module in the second stage the interaction modeling performance is significantly improved.

\section{The ICNN network}
\label{sup:icnn}

For the IRs equal to 80m, 60m, 40m, 20m, and 5m, we set the grid sizes of the feature map crops to 64, 48, 32, 16 and 4, respectively. 
Zero-valued padding is utilized in the convolutional layers when necessary.

\begin{figure*}[!htbp]
    \centering
    \includegraphics[width=0.32\textwidth]{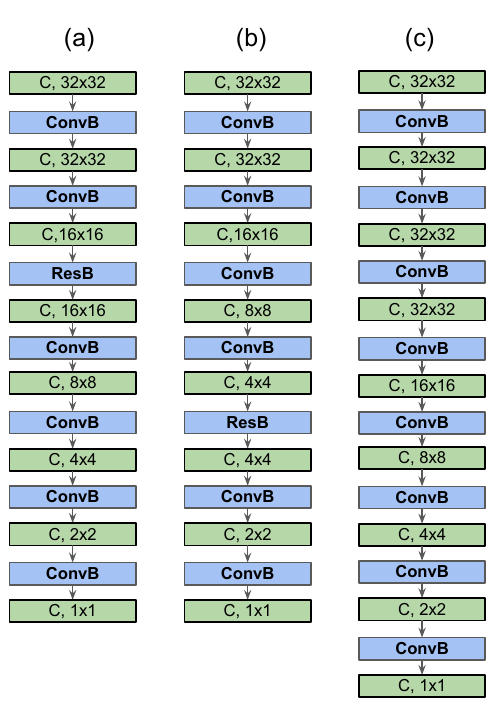}
    \caption{Various considered designs for ICNN, using 32$\times$32 input as an example. The green boxes $(C, W \times W)$ represent tensors of grid size equal to $W \times W$ and $C$ channels ($C$ is equal to 256 in all designs). The presented main results were based on design (a).}
    \label{fig:icnn}
\vspace{0.2cm}
\end{figure*}

We did not extensively investigate network designs for the ICNN module. 
Several straightforward options (see Fig. \ref{fig:icnn}) that stacked {\bf ConvB} and {\bf ResB} blocks in series were evaluated empirically.
These options set the strides of the last few {\bf ConvB} blocks to 2 so that the input feature map crop was down-sampled gradually to $1\times1$ after being processed by the ICNN. 
We observed that the model performance was not sensitive to the changes in these ICNN variants.

\section{The GNN network}
\label{sup:gnn}
Two-layer MLPs are used in a few places in the GNN networks of this work. All output vectors through the MLPs have the identical dimension as $fv$, 256. In other words, the dimensions remain unchanged through the MLPs.

Note that both max-pooling and mean-pooling were studied in the GNNs (see Eq.~\ref{eq:pool}) and no experimental difference was observed.
We also explored adding other relative relations such as relative velocities to the graph edge attribute and observed insignificant changes to the model performances. 
Besides, in the main text all node and edge attributes were based on deterministic model outputs. 
We studied using probabilistic (Gaussian and Laplace) outputs for the attributes, and only measured performance difference within the metric variance level. 
Note that the GNNs experimented in this work built edges between all vehicle actors. 
It was clear that some of the edges were unnecessary (e.g, between two far-apart vehicles that were driving farther apart). 
However, given the high vehicle speed and long prediction horizon of $4$s, such optimization was not straightforward and thus not experimented with in more depth.

\section{The training setup}
\label{sup:setting}
Each training sequential example comprises 10 past and current sweeps ($-0.9$s, $-0.8$s, \ldots, $0$s), and 41 current and future timestamps for ground-truth supervision ($0$s, $0.1$s, \ldots, $4.0$s). 
The frame at current timestamp is referred to as the key frame. 
Each scene on the in-house data set is 25s long, producing at most 200 complete sequential examples. 
We trained all of the models with decimated key frames in the training split once (i.e., every sequential example whose key frame is at $t$, $t+0.2$s, $t+0.4$s, $\ldots$, is used once during model training).

The models were implemented in PyTorch \cite{NEURIPS2019_9015} and trained end-to-end with 16~GPUs (Nvidia RTX 2080), with a batch size of 2 per-GPU. 
Training without the GNN module is completed in about $12$ hours. 
We use the Adam optimizer \cite{kingma2014adam} with a learning rate of $2$e-$4$, decayed to $2$e-$5$ at $75\%$ and $2$e-$6$ at $95\%$ of the training iterations.

For the models with the interaction loss, the weight of the interaction loss was set to $2.0$, which slightly outperformed those with weights of $1.0$ and $3.0$ in the interaction metrics. No significant displacement error difference was observed.  

\section{The data}
\label{sup:data}
In the Methodology Section, we briefly explained that large data was required to conduct experiments with low metric variances and derive general conclusion. Here we provide more details of the in-house data set and its comparison to some open-sourced data sets.

As shown by the experimental results, the overlap rates are low, particularly for models equipped with high abilities in modeling interaction. This means large and diverse test data is required to achieve low metric variances. Take the popular nuScenes data set~\cite{caesar2020nuscenes} for autonomous driving as an example, its training, validation, and test sets for prediction task combined have 1000 scenes (each is 20s long). 
On the in-house data, our work uses a test set of 5000 scenes (each is 25s long), which can help facilitate the comparison and analysis of fine differences (e.g., those between {\bf +GNN} and {\bf +GNN (no edges)} at large IRs, see metric variances in the next section).

\begin{figure*}[ht!]
    \centering
    \includegraphics[width=0.6\linewidth]{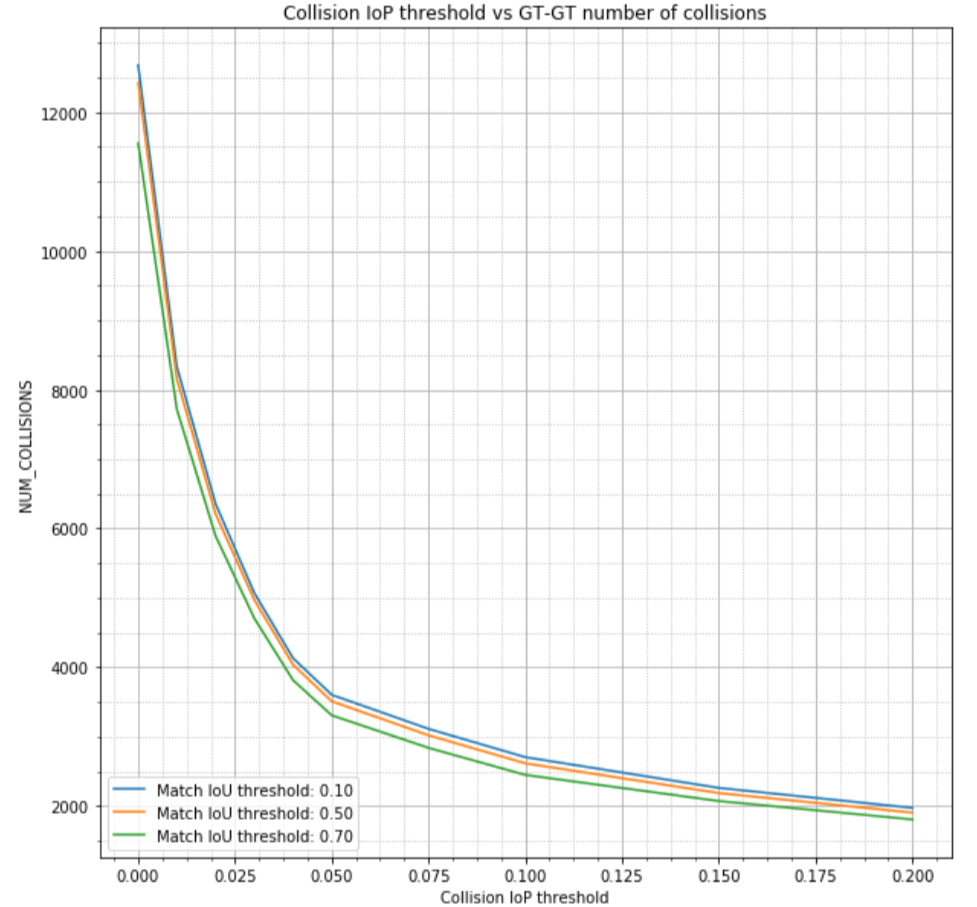}
    \caption{Number of ground-truth label overlaps of all timestamps vs. IoP through the test set. Here we only consider label trajectories whose boxes at the key-frame match the detections of the baseline model. We threshold the matching in terms of IoU at 0.1, 0.5, and 0.75 on the plot. The number of overlaps is non-zero even at high IoP,  because large overlaps indeed occur in this data set, for instance, when the arm of a construction vehicle is over another vehicle, which would be a large overlap in the bird's eye view. }
    \label{fig:gt_collision}
\end{figure*}

In pure trajectory prediction tasks where the metric variance of displacement error is low even based on smaller data sets, we observed that the methodological comparison concluded on this in-house data set was generalized to other AV data sets. Specifically, in our earlier trajectory prediction works ~\cite{djuric2020multixnet, su2020temporallycontinuous}  we found the results on this in-house data set to correlate with the smaller open-sourced data sets. Comparing the trajectory prediction performance on the nuScenes data, it was also observed that the $3$s displacement errors (vehicles) were $1.58$, $1.45$, and $1.04$m for CAR-Net~\cite{sadeghian2017car-net, casas2019spatially}, SpaGNN~\cite{casas2019spatially}, and our conv-only model (+ICM (IR=60m)), respectively. For fair comparison, all models were measured at a same detection recall of $60\%$.

Finally, we confirm that the observation that the convolutional approach performed comparably to or better than GNN remained valid with smaller data subsets. E.g., trained with 1/3 of the full training set (training time is also 1/3 long) and evaluated on the same large test set (5000 scenes), we measured displacement errors of 0.818, 0.641, and 0.714m in the baseline, +ICM (IR=80m), and +GNN (IR=0m) models, respectively. Their actor-actor overlap rates were 3.26, 1.04, and 1.70$\%$; actor-static overlap rates were 1.22, 0.49, and 0.66$\%$, respectively. These results suggest that the empirical studies presented in this work are generally valid.

We complete the discussion on data with two final notes. 
The first note is about the labeling standard for object detection. Actors that were far away from the traffic regions (e.g., roads and  side-walks) were excluded during training and performance evaluation, which could lead to a higher detection performance than those based on data sets that considered all actors in the scene. The second note is about the distribution of the multimodal predictions, where approximately $80\%$ of the moving vehicles were in the ``straight''-driving mode (as opposed to left and right turns).

\section{The metrics}
\label{sup:metric}

The overlap interaction metrics are defined in terms of  the ratio of intersection between two objects to the the area of the smaller object (IoP) rather than the more common IoU, because the latter is insensitive to an overlap between a large object and a small object. Although there are no collisions between actors in the data set, we have measured overlap rates between actor labels because some labeling boxes are slightly and imprecisely over-sized (thus the learnt object detections would be over-sized too). In Figs.~\ref{fig:gt_collision} we plot the number of overlaps between label actors as a function of IoP threshold. The number drops rapidly after IoP at $0.05$ approximately, meaning thresholding IoP at $0.05$ would just eliminate the majority of the false positive overlaps. We adapted this threshold value in evaluating the overlaps of predicted trajectories. Such setting allowed the metrics to measure interaction modeling in the experiments robustly and indicatively.

The proposed convolutional approach can be applied to improve interaction modeling for other traffic participants such as pedestrians and cyclists. We observed significant effects in case studies. The studies were not discussed in this work because even labels of pedestrians and cyclists often had considerable and arguably correct overlaps. As a result, the overlap rates were no longer unambiguous metrics for interaction modeling. We thus limited our discussion to interaction between vehicles and vehicles, and between vehicles and static obstacles. 

By including these careful designs for the data and metrics, the resulting metrics variances were 0.004m, 2e-4, and 3e-4 for DE, actor-actor, and actor-static collision rates, respectively, computed by training {\bf +ICM} (80m) 4 times (other models were trained and evaluated once). This level of variances allowed us to resolve difference in interaction modeling between models, even on rare events such as overlaps.

\section{Additional results focusing on overlaps with non-vehicle actors}
\label{sup:non-veh}
\begin{figure*}[ht!]
    \centering
    \includegraphics[width=0.32\textwidth]{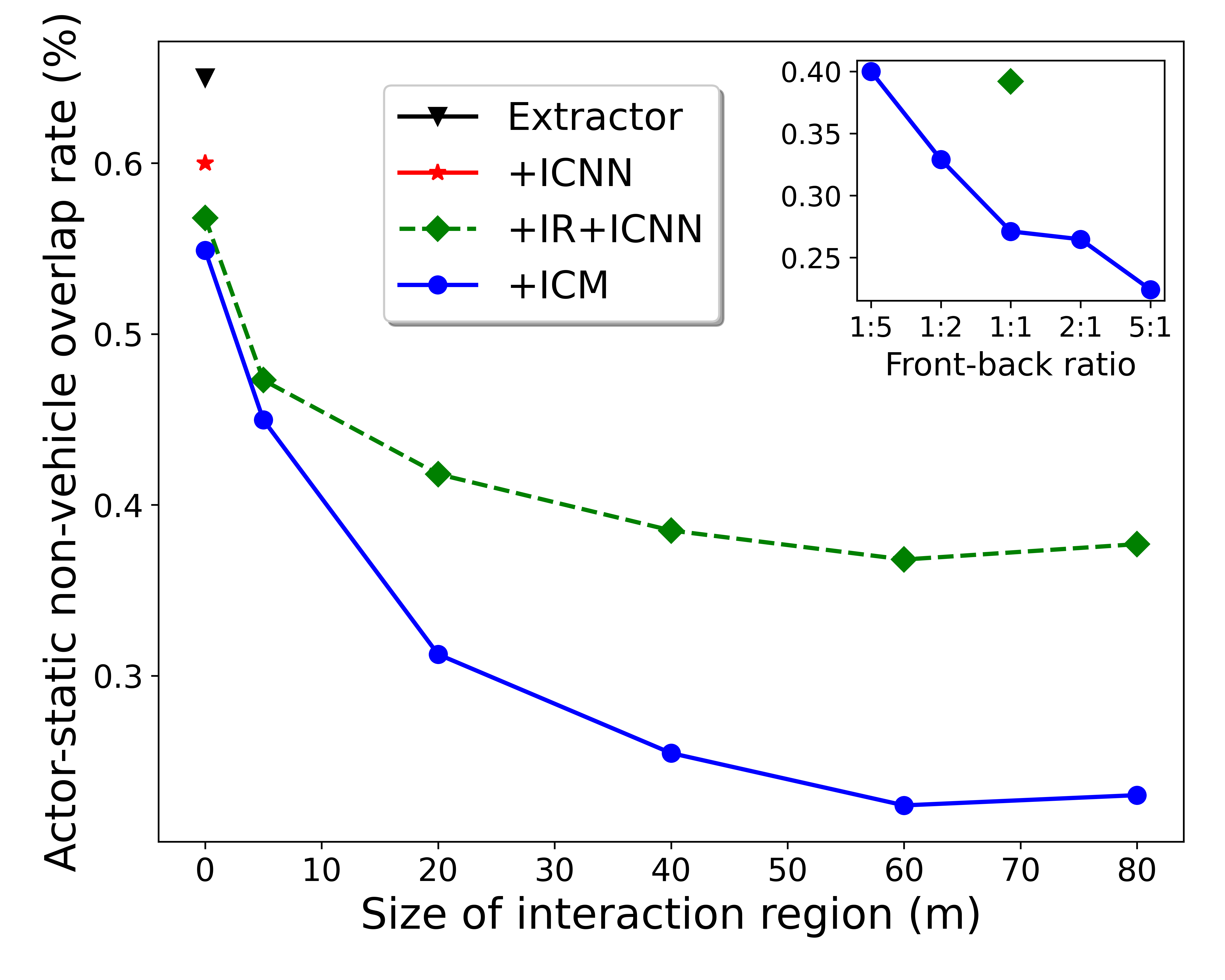}
    \includegraphics[width=0.32\textwidth]{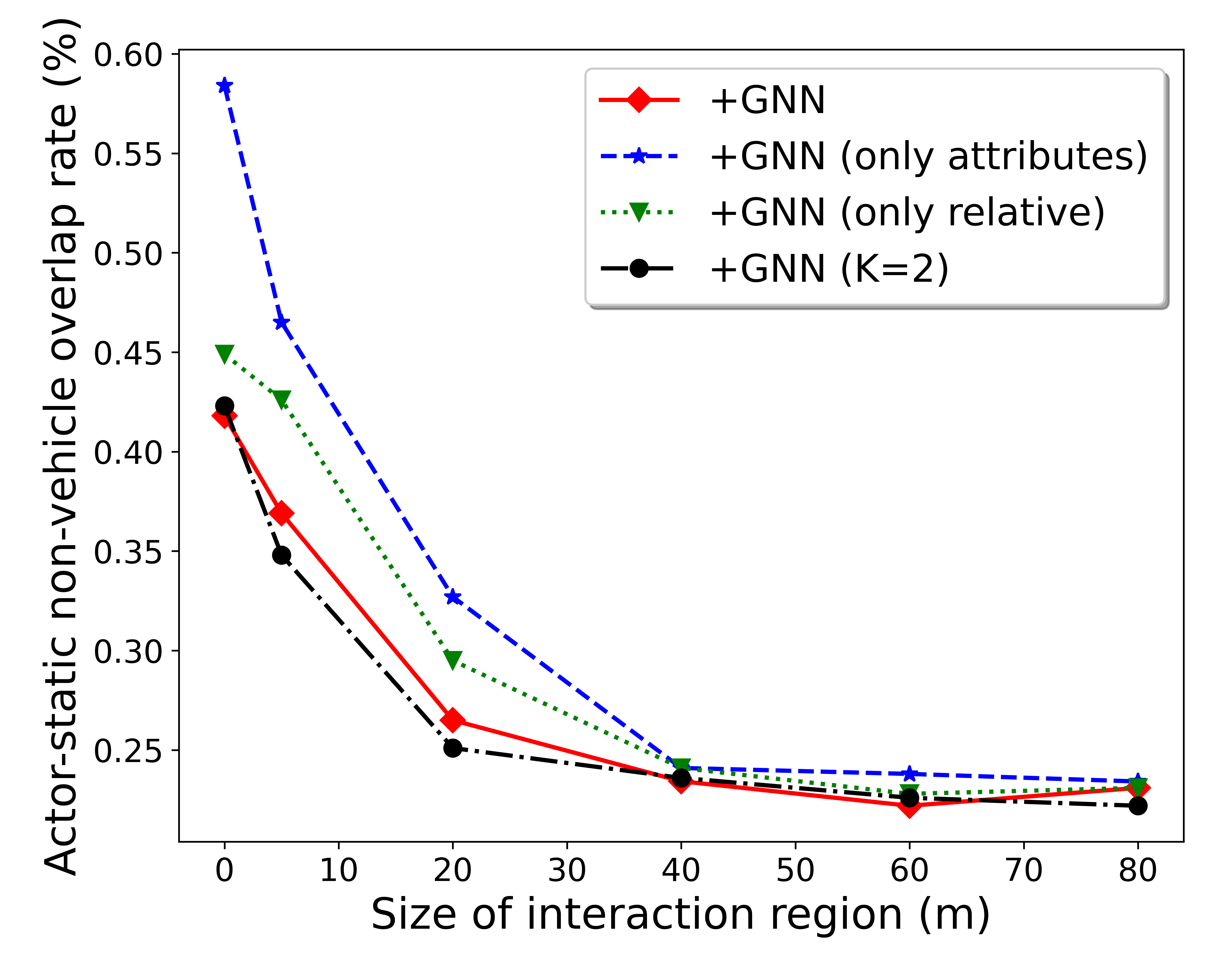}
    \includegraphics[width=0.32\textwidth]{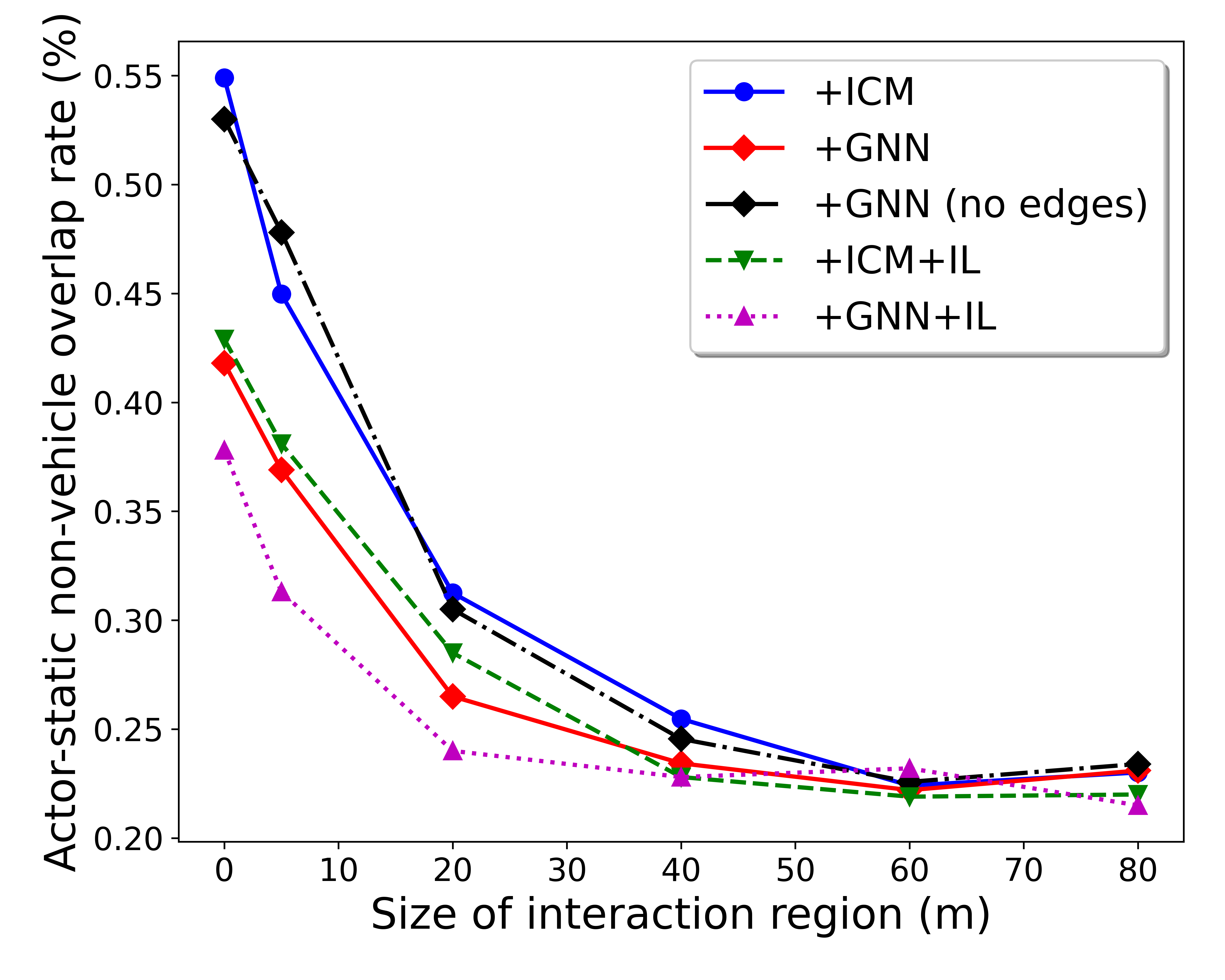}
    \caption{Overlap rate of forecasted actor trajectories overlapping with static {\it non-vehicle} traffic objects.}
    \label{fig:metrics_nonveh}
\end{figure*}

The actor-static overlap rate in the main paper considers overlaps between forecasted trajectories with both vehicle and static non-vehicle traffic objects.
In this section, we provide additional results focusing on overlap with static {\it non-vehicle} traffic objects. 
Here the overlap rate is defined as the percentage of forecasted trajectories of detected actors that overlap with ground-truth static {\it non-vehicle} traffic objects. 
The three panels in Fig. \ref{fig:metrics_nonveh} correspond to Figs.~\ref{fig:ICM}
-~\ref{fig:compare} in the main paper. 
As the feature map input cropped by the IR covers features of both vehicle and non-vehicle traffic objects in the ICM approach, it is not surprising that ICM effectively improves this interaction metric too. 
It is, however, interesting to note that even though GNN does not build nodes for the non-vehicle traffic objects in the graph, it also lowers this overlap rate by $24\%$, by comparing {\bf +ICM} (0m) to {\bf +GNN} (0m). 
The reduction is attributed to the fact that by avoiding overlaps with vehicles (after adding the GNN), the overlaps with some of the non-vehicle objects near those vehicles are also avoided. 
Another factor may be the proximity effect of CNNs, as the pixel features of the vehicle actors might comprise information about its nearby non-vehicle objects. 
The improvement of GNN on the overlap avoidance with non-vehicle objects ($24\%$), however, is considerably lower than that with vehicle actors ($42\%$ as shown in the main paper by comparing {\bf +ICM} (0m) to {\bf +GNN} (0m) in Fig.~\ref{fig:compare} right), which is reasonable as the GNN does not model the interactions with non-vehicle objects directly.

\section{Videos}
\label{sup:video}
In addition to Fig.~\ref{fig:obstacle}, we provide qualitative results with three \href{https://youtube.com/playlist?list=PLbI8u9Kk9gFyWIP7T9aWWvoO6nrEUAs1W}{videos}\footnote{https://youtube.com/playlist?list=PLbI8u9Kk9gFyWIP7T9aWWvoO6nrEUAs1W} where the predictions of the baseline ({\bf +ICM}, 0m) are on the left and the predictions of the ICM model ({\bf +ICM}, 60m) are on the right. The overlapped obstacles are filled in red, the ground-truth trajectories are in grey, and the forecasted trajectories are in blue. Trajectory visualization is downsampled to $2$Hz for clarity. Different from Fig.~\ref{fig:obstacle}, we visualize the predictions of all actors in the common AV frame of reference. Note that the videos are 20s long, because each scene in the data set is 25s long, where the first second is used for the 10-sweep input and the last four seconds are used for the 4s forecasting time horizon.
\end{appendices}

\end{document}